\documentclass[conference]{IEEEtran}
\IEEEoverridecommandlockouts
\usepackage{cite}
\usepackage{amsmath,amssymb,amsfonts}
\usepackage{algorithmic}
\usepackage{graphicx}
\usepackage{textcomp}
\usepackage{xcolor}

\usepackage{float}
\usepackage{balance}
\usepackage{makecell}
  
\usepackage{bigstrut}
\usepackage{array}
\usepackage{graphicx}
\usepackage{subfigure}
\usepackage{amssymb}
\usepackage[algo2e, ruled, lined, linesnumbered]{algorithm2e}
\usepackage{multicol,lipsum}
\usepackage{cuted, nccmath}
\usepackage{colortbl}
\usepackage{multirow}
\SetAlFnt{\small}
\SetAlCapFnt{\small}
\SetArgSty{textnormal}
\usepackage{xcolor}
\usepackage[T1]{fontenc}
\usepackage[utf8]{inputenc}
\usepackage[font=footnotesize,labelfont=bf]{caption}
\usepackage{booktabs}

\newtheorem{example}{Example}
\newtheorem{definition}{Definition}

\def\BibTeX{{\rm B\kern-.05em{\sc i\kern-.025em b}\kern-.08em
    T\kern-.1667em\lower.7ex\hbox{E}\kern-.125emX}}
\begin{document}

\title{Parsimonious Morpheme Segmentation with an Application to Enriching Word Embeddings}

\author{\IEEEauthorblockN{$^1$Ahmed El-Kishky, $^2$Frank Xu, $^3$Aston Zhang, $^1$Jiawei Han}
\IEEEauthorblockA{$^1$The University of Illinois at Urbana-Champaign, $^2$Carnegie Mellon University, $^3$Amazon AI \\
 \{elkishk2,hanj\}@illinois.edu, frankxu@cmu.edu, astonz@amazon.com }
}

\IEEEoverridecommandlockouts
\IEEEpubid{\makebox[\columnwidth]{978-1-7281-0858-2/19/\$31.00~\copyright2019 IEEE \hfill} \hspace{\columnsep}\makebox[\columnwidth]{ }}
\maketitle
\IEEEpubidadjcol

\begin{abstract}
Traditionally, many text-mining tasks treat individual word-tokens as the finest meaningful semantic granularity.  However, in many languages and specialized corpora, words are composed by concatenating semantically meaningful subword structures. Word-level analysis cannot leverage the semantic information present in such subword structures. With regard to word embedding techniques, this leads to not only poor embeddings for infrequent words in long-tailed text corpora but also weak capabilities for handling out-of-vocabulary words. In this paper we propose MorphMine for unsupervised morpheme segmentation. MorphMine applies a parsimony criterion to hierarchically segment words into the fewest number of morphemes at each level of the hierarchy. This leads to longer shared morphemes at each level of segmentation. Experiments show that MorphMine segments words in a variety of languages into human-verified morphemes. Additionally, we experimentally demonstrate that utilizing MorphMine morphemes to enrich word embeddings consistently improves embedding quality on a variety of of embedding evaluations and a downstream language modeling task.
\end{abstract}


\section{Introduction}
Decomposing individual words into finer-granularity morphemes is a necessary step for automatically preprocessing concatenative vocabularies where the number of unique word forms is very large. While linguistic approaches can be used to tackle such segmentation, such rule-based approaches are often tailored to specific languages or domains. As such, data-driven, unsupervised methods that forgo linguistic knowledge have been studied \cite{hammarstrom2011unsupervised,creutz2002unsupervised}. Typically, these methods focus on segmenting words by applying a probabilistic model or compression algorithms to a full text corpus. The resultant morphemes from these methods have been primarily shown to improve neural machine translation~\cite{sennrichneural,kudo2018subword}.

One natural application to utilize these semantically meaningful morphemes is distributed word representation. There are many advantages to using distributed continuous word representations as an alternative to one-hot bag of words~\cite{rumelhart1988learning,elman1990finding} since this leads to a dimensionality much smaller than the vocabulary size of a corpus. It has been shown that working with low-dimensional representations not only demonstrates computational efficiency, but also captures syntactic and semantic regularities while boosting the performance in text classification, sequential classification, sentiment analysis, and machine translation~\cite{mikolov2013linguistic,joulin2017bag,huang2015bidirectional,tang2014learning,zou2013bilingual}. As such, many methods have been developed to learn these word representations from large, unlabeled text corpora~\cite{collobert2008unified,mikolov2013efficient,mikolov2013distributed}.

Despite many advances, unsupervised learning of distributed representations can struggle in learning adequate vectors for infrequent words. This problem is ubiquitous because most text corpora demonstrate long-tail distributions in relation to word frequency, with often $40\%-60\%$ of words in a vocabulary appearing just once in a corpus~\cite{kornai2007mathematical}. Naturally, many methods fail to produce meaningful embeddings for unseen (out-of-vocabulary) words. Using morphemes for parameter sharing not only bolster training data for infrequent words but also allow for constructing meaningful word embeddings for unseen words.

\begin{center}
 \begin{figure}[t]
\centering
\includegraphics[scale=0.35]{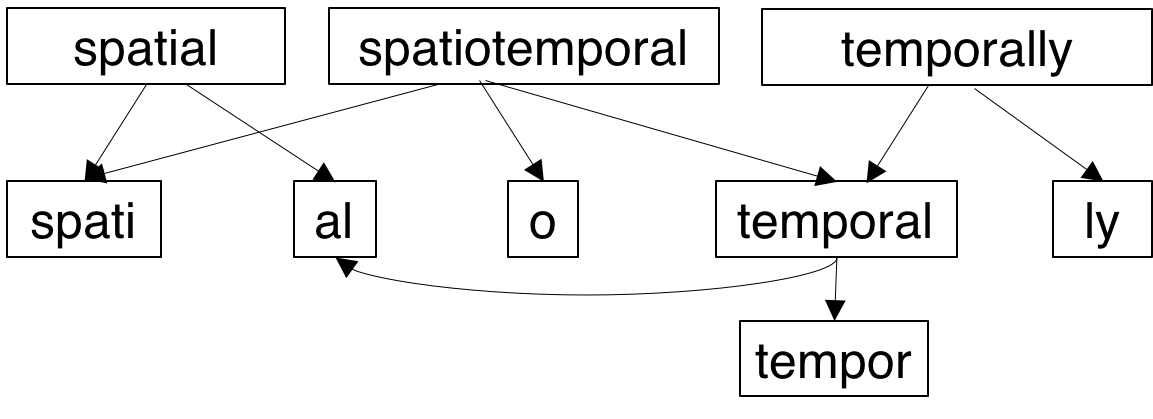}
\caption{\small{Hierarchical segmentation of words.}}
\label{fig:paramsharing}
\end{figure}
\end{center}

What differentiates our method from others is \textit{extracting morphemes at multiple granularity}. As seen in Figure~\ref{fig:paramsharing}, morphologically-rich words share semantically-meaningful morphemes. Larger morphemes carry more semantic meaning, but are often infrequent within the vocabulary and discarded in favor of more frequent finer-grained morphemes in other methods. Yet, including both fine and coarse-grained morphemes, can better semantically tie the meanings of words that share them. With this motivation, we propose MorphMine, a continuation on preliminary work~\cite{el2018entropy}. We formalize the novel methodology by framing the morpheme segmentation as entropy-boundary identification and segmentation with a parsimony criterion. We introduce a global resegmentation to refine and improve the segmentation after the initial segmentation. Finally, we evaluate our method on a variety of datasets and tasks in multiple language and demonstrate how multi-granular morphemes can be used for enriching word embeddings for robustness to data-sparsity. 

\section{Preliminaries}
\label{sec:subwordminingframework}
The input is a corpus $W$, consisting of $|W|$ words: $W = w_1,\ldots,w_{|W|}$. From this corpus, we construct a vocabulary of unique words, $V$, of size $|V|$ such that $\forall w\in W, w\in V$. 
In addition, the $v^{th}$ word is a sequence of $|v|$ characters: $c_{v,i}, i=1,\dots,|v|$. 
For convenience we index all the unique characters that compose the input vocabulary with $C$ characters and $c_{v,i} = x, where ~x\in\{1,\dots,C\}$ means that the $i^{th}$ character in $v^{th}$ word is the $x^{th}$ character in the character vocabulary.

Given an input corpus consisting of a word sequence and a vocabulary list of unique words, our goal is to segment the vocabulary list to identify human-interpretable and semantically meaningful morphemes, then utilize these morphemes for parameter sharing when learning distributed word representations from the corpus. 

\begin{definition}[Morpheme Formalization]
\label{def:subword_formalization}
\begin{itemize}
	\item A \emph{morpheme} is a sequence of characters: $m$ = $\{c_{v,i}, ..., c_{v,i+n}\}\; ~where~ n> 0$
	\item A \emph{partition} over vocabulary word $v$ is a sequence of morphemes: $\mathcal{G}_v$ = $(m_{v,1}, \dots, m_{v,G_v} ) \;~where ~G_v\geq 1$ s.t. the concatenation of the morphemes is the original word.
\end{itemize}
\end{definition}
In Definition~\ref{def:subword_formalization} we formalize a morpheme and the resultant partition from segmenting a word into morphemes. In addition we outline the desired properties of the framework as follows: 
\begin{enumerate}
    \item extracts semantically meaningful, human-interpretable at multiple granularity
    \item the method is general and applies to words on a variety of languages
    \item enriching word morphemes improves word embeddings
    \item the overall method is computationally efficient
\end{enumerate}

\subsection{The MorphMine Framework}
At a high-level, our proposed framework can be summarized into two sequential steps: (1) mining candidate morpheme patterns and character co-occurence statistics, and (2) performing word segmentation into finer-grained morphemes. 
In step one, by applying an information-theoretic metric to detect candidate morpheme boundaries, we identify candidate morphemes within each vocabulary word. These morphemes are propagated to other words and pruned to ensure high-quality. For step two, from this candidate pool, we then apply an unsupervised dynamic programming segmentation algorithm to select a subset of these morphemes that best segment each word. Segmentation and partition induction further prune away low-quality morpheme candidates leaving a high-quality morpheme vocabulary. After inducing a partition on each word, we can recursively segment each morpheme to finer granularity. Applying this two-step process maps each word in the input vocabulary to a set of high-quality morphemes. The resultant morphemes from the hierarchical segmentation can then be used for downstream NLP and text analysis tasks.

The main objective in morpheme pattern mining is to collect aggregate statistics on morpheme patterns that can be used to score and reason about the quality of candidate morphemes. These statistics are then used in the word segmentation algorithm. 
For each character $n$-gram that appears more than once in the vocabulary, there is a potential for parameter sharing via the candidate morpheme as it appears in multiple vocabulary words. 
Additionally the frequency counts of these morphemes will be used for entropy-boundary computation to identify and score potential morpheme candidates. 
These candidates are input to the word-segmentation algorithm that attempts to apply Occam's Razor by positing that using the fewest morphemes in the segmentation best segments each word~\cite{gauch2003scientific}. This process is then applied recursively to each morpheme to obtain finer-grained morphemes.

By inducing a partition over each vocabulary word, we effectively transform each word into a bag-of-morphemes. These morphemes can be shared among other words within the vocabulary and model the belief that words that share morphemes, share semantic meaning. This is done by individually embedding each morpheme; these morpheme embeddings are then combined to form the final word embedding. Because a word embedding is constructed from the embeddings of constituent morphemes, words that share constituent morphemes will be \textit{partially} constructed from similar morpheme embeddings.

We expound upon our morpheme-mining algorithm and its evaluation in a popular embedding framework in Section~\ref{sec:segmenting}.

\section{Methodology}
\label{sec:segmenting}
Given an input vocabulary list $V$, MorphMine segments each word into non-overlapping character $n$-grams (morphemes). 
Our method is non-parametric, hierarchical and data-driven, allowing for good cross-domain performance without incorporating domain-specific knowledge or linguistic rulesets. 
The entire morpheme segmentation can be performed as an easy preprocessing step to the vocabulary for downstream text-related tasks. To learn a morpheme vocabulary and segment an input vocabulary, MorphMine performs the following steps: (1) mine morpheme pattern counts and compute entropy statistics, (2) apply parsimonious segmentation to identify the best \textit{locally-consistent} segmentation, (3) recompute morpheme counts after segmentation to ensure \textit{global-consistency} and maximize parameter sharing of morphemes, and (4) re-segment using refined morpheme vocabulary counts.

We apply an entropy-based scoring function to identify morpheme boundaries: generating candidate morpheme vocabulary. Given this collection of morphemes and their counts, the next step is to apply a dynamic-programming algorithm to segment each word into high-quality morphemes. For each word, the parsimonious segmentation identifies the \textit{most-likely} segmentation using the \textit{fewest number of morphemes}. This step discards a large number of lower quality candidates morphemes from our vocabulary and allows for a more-accurate estimate of morpheme counts. Using the refined vocabulary, we can then re-segment and improve the overall quality of segmentation. The re-segmentation biases towards selecting locally-consistent segmentations that globally-optimize for morpheme parameter sharing. That is, the resegmentation favors morphemes used in the segmentation of other words in the vocabulary. Finally, the resultant collection of morphemes for each word can be utilized to enrich word embeddings.

\subsection{Morpheme Vocabulary Generation}

\label{subsec:candidate}
Our segmentation of words into morphemes relies on the idea of morpheme compositionality. 
That is, the input vocabulary can be constructed by composing morphemes drawn from a smaller morpheme vocabulary.  As such we introduce an approach for creating the initial morpheme vocabulary: prefix, suffix, and root-word candidates.

\subsubsection{Prefix \& Suffix Generation.}
We posit that prefixes and suffixes can be identified through the concept of transition predictability.

\vspace{0.2cm}
\begin{definition}[Transition Predictability]
 \label{def:predictability}
 Transition predictability is a quantification of being able to predict the next character in a word given a prefix.
\end{definition}
\vspace{0.2cm}

Previous works have attempted to quantify Definition~\ref{def:predictability}, by using number of character choices following a prefix in a vocabulary~\cite{harris1970phoneme,hafer1974word,dejean1998morphemes,saffran1996word}. For example, many words begin with the prefix, ``\textit{pre}'' such as, \textit{prepaid}, \textit{preview}, \textit{presoak}, etc. Given the large number of words with the prefix ``pre'', the transition from ``pr'' to ``pre'' predictable, but ``pre'' to a longer prefix is not as predictable as many words have ``pre'' followed by a variety of root words.

Unfortunately, using raw counts to identify high-unpredictability boundaries for prefixes does not generalize to large vocabularies and different languages as the character count is arbitrary. As such we propose a metric on the normalized distribution of character choices: information entropy~\cite{shannon2001mathematical}. Let $v$ be a word consisting of $|v|$ characters and $m_i$ be a prefix of $v$ ending at the $i_{th}$ character of $v$. 
For each candidate prefix boundary $i$ for $i \in [1 \ldots |v| ]$, the prefix transition unpredictability can be quantified with information entropy. As the transition between a prefix and longer prefixes can be modeled as a multinomial of support size $C$, the character vocabulary, we use the multinomial distribution entropy:

\begin{equation*}
\begin{aligned}
H(X)& = -\log(n!) - n\sum_{j=1}^{C} p_j \log(p_j) + \\
&~~~~\sum_{j=1}^{C} \sum_{x_j=0}^{n} {n\choose x_j}p_j^{x_j}(1-p_j)^{n-x_j}\log(x_j!) \\
&= -\sum_{j=1}^C p_j \log( p_j), \text{ when n=1}
\end{aligned}
\end{equation*}
This is the entropy of a multinomial over support $C$ for $n$ independent trials each of which leads to a success for exactly one of the $C$ characters. 
Because we consider a single trial, $n=1$, we simplify it into entropy of the categorical distribution. 
For the prefix $m_i$:
\begin{equation*}
H(m_i) =  -\sum_{j=1}^C \mathsf{P}(m_{i} \oplus c_j | m_i) \times\text{log}_2 \mathsf{P}(m_{i} \oplus c_j | m_i),
\end{equation*}
where $\oplus$ denotes the binary string concatenation of two strings and the transitional prefix probability is estimated as:
\begin{equation*}
\mathsf{P}(m_{i} \oplus c_j | m_i) = \frac{f(m_{i} \oplus c_j)}{f(m_{i})}
\end{equation*} 
and $f(m_{i})$ denotes the frequency of a prefix $m_{i}$ in the input vocabulary list. 
The entropy of suffixes can, without loss of generality, be similarly computed by reversing each word in the vocabulary and treating each suffix as a prefix.
\begin{center}
\begin{figure}
     \centering
     \includegraphics[scale=0.31]{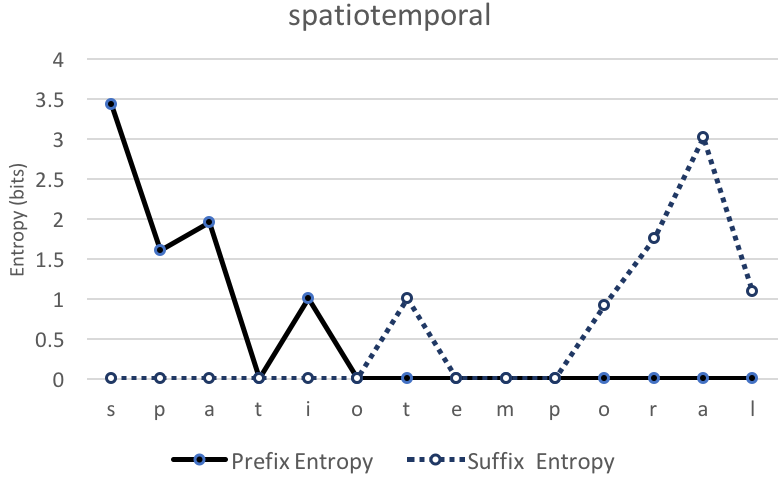}
     \caption{Prefix and suffix transition entropy.}
    \label{fig:entropy}
\end{figure}
\end{center}

The information entropy of each possible prefix and suffix in the vocabulary is computed in \textit{linear time} with relation to unique vocabulary size using a prefix tree data structure to store counts over prefixes. Given entropy scores for each prefix and suffix, scores are computed for each candidate split point in each word. Under the entropy scoring of prefixes and suffixes, we identify \textit{local maxima} in entropy as candidate boundaries for prefixes and suffixes. That is entropy of a prefix one-character shorter and one-character longer should be lower than a candidate prefix boundary. This is intuitive as under our principle of compositionality assumption, complex words are formed by concatenating morpheme structures. As such, given an incomplete morpheme, the next character can easily be predicted, but given a complete morpheme, any number of new morphemes can be concatenated to the completed morpheme increasing the unpredictability and thus entropy. These high-entropy positions thus serve as a strong indicator of morpheme boundaries.  As seen in Figure~\ref{fig:entropy}, for the word ``spatiotemporal'',  candidate prefixes and suffixes are found at boundaries exhibit a local maxima in entropy. For ``spatiotemporal'', candidate prefixes are ``spa'' and ``spati'' while candidate  suffixes include ``al''  and ``temporal''. 

\subsubsection{Root Word Generation}
Utilizing entropy-scoring, it is possible to detect morpheme structures that occur at the beginning or end of a word. However, many words often contain morpheme structure between prefixes and suffixes. 
For each prefix and suffix candidate identified in a word, it is possible to generate many candidate root words by stemming the word and removing prefixes and suffixes. This creates a high-quality pool of root words to be used in conjunction with prefixes and suffixes for segmenting the vocabulary.

\begin{example}[Root Extraction]
\label{eg:root}
Removing prefixes and suffixes yields candidate roots.
\begin{center}
\textbf{[pre]} + \underline{authenticat} + \textbf{[ion]}\\
\textbf{[pre]} + \underline{authentication}\\
 \end{center}
 The characters grouped together by \textbf{[]} are prefixes and suffixes. When removed, the remaining underlined character-sequence represent candidate root words. 
\end{example}
\vspace{0.1cm}

As seen in Example~\ref{eg:root}, when stripping the combinations of prefixes and suffixes of a word, the remaining character sequence is considered a candidate root word. We apply some filtering conditions for each candidate root to test the viability as a shareable root. These include: (1) a minimum support of two within the vocabulary, and (2) the minimum root length of four. Additionally, for each word in the vocabulary, after stripping prefixes and suffixes, the candidate root words that meet the constraints are added to the morpheme vocabulary.

\subsection{Parsimonious Morpheme Segmentation}
\label{subsec:segmenting_words}
After generating a morpheme vocabulary using entropy-based predictability metric for boundary detection, we segment words into morphemes, utilizing this morpheme vocabulary. The algorithm first identifies candidate morphemes from the morpheme vocabulary within a word, then selects a subset of these candidate morphemes that best segment the word. The main insight is a per-word implementation of Occam's Razor. That is, according to the preference for parsimonious hypotheses, we posit that each word is composed of the \textit{fewest number of morphemes that maximally cover the word}.

\begin{center}
\begin{figure}
    \subfigure[Parsimonious Segmentation]{
        \label{fig:rvd-stat-1}
        \begin{minipage}{0.24\textwidth}
         \centering
         \includegraphics[height=2.5cm]{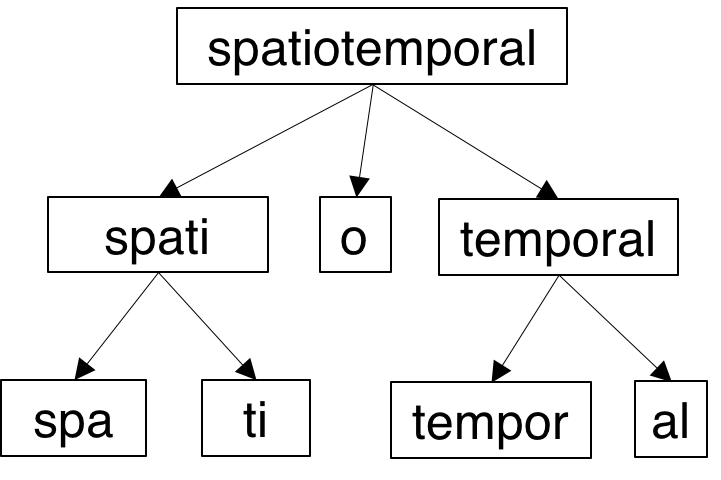}
        \end{minipage}\hfill
    }
    \subfigure[Candidate Morphemes]{
        \label{fig:rvd-stat-2}
        \begin {minipage}{0.24\textwidth}
         \centering
         \includegraphics[height=2.5cm]{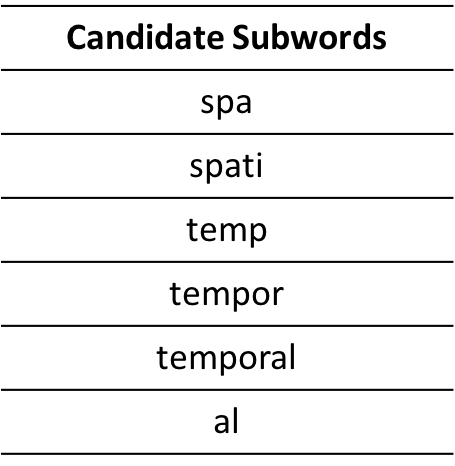}
        \end{minipage}
    }
    \caption{Segmentation of the word ``spatiotemporal'' using disjoint interval covering.}
    \label{fig:segment_parsimony}
\end{figure}
\end{center}

As seen in Figure~\ref{fig:segment_parsimony}, morphemes present in the target word are identified and recursive segmentation is performed to segment the word into morphemes. Example~\ref{eg:parsimony} demonstrates how the candidates are used to segment the target word under the \textit{parsimony criterion}. 

\begin{example}[Parsimonious Segmentation]
\label{eg:parsimony}
Segmentations are scored based on word coverage and the number of morphemes.
\begin{center}
\begin{tabular}{ lll}
  \hline
 \textbf{\emph{Segmentation}} & \textbf{\emph{\# Morphs}}  & \textbf{\emph{Coverage}}\\
  \hline
  \hline
  {[\textbf{spa}]} +  tio+ {[\textbf{temporal}]} & 2 & 11 \\
  \hline
\rowcolor{gray!30}{[\textbf{spati}]} + o + {[\textbf{temporal}]} & 2 & 13 \\
\hline
  {[\textbf{spati}]} + o + {[\textbf{tempor}]} + {[\textbf{al}]}& 3 & 13 \\
  \hline
  \hline
    {[\textbf{spa]}} + tio {[\textbf{tempor}]} + {[\textbf{al}]} & 3 & 11 \\
  \hline
\end{tabular}
\end{center}
\vspace{0.2cm}
The highlighted row displays the maximally parsimonious morpheme segmentation.
\end{example}
\vspace{0.1cm}

Subsets of non-overlapping candidate morphemes are used in segmentation, and the most parsimonious segmentation is selected. Because the possible subsets of candidate morphemes form a power set, direct enumeration of each segmentation quickly proves computationally slow for even a modest number of candidate morphemes. To identify the most parsimonious segmentation, we abstract our parsimonious morpheme segmentation task into a general problem we dub \textit{Disjoint Interval Covering} and demonstrate that this problem can be solved via dynamic programming in linear time. 
We formalize the disjoint interval covering problem as follows:
\vspace{0.2cm}
\begin{definition}[Disjoint Interval Covering]
\label{def:interval_covering}
Given an input $N\in \mathbb{N}$ and a set $A$ of pairs $(a,b): a, b \in \{1 \ldots N\} \times \{1 \ldots N\}$ and $a < b$, find the smallest subset $B\subseteq A$ such that $|\underset{x \in B}{{\bigcup x|}}$ is maximized, $|B|$ is minimized, and $\forall x,y \in B: x\neq y \Rightarrow x \cap y = \varnothing$.
\end{definition}
\vspace{0.2cm}

As seen in Definition~\ref{def:interval_covering}, the input is a set of pairs $A$ and a positive integer $N$. Within the segmentation perspective, these refer to position index boundary pairs for candidate morphemes and the word length. Given these inputs, the objective is to select a minimum subset of disjoint morphemes that maximally cover the word. That is, select a set of disjoint morpheme whose combined length is as close as possible to the word length.

\begin{scriptsize}
\begin{equation}
\label{eq:1}
    F(j) = \max_0 \min_1 \left\{\begin{array}{lr}
        (0, 0), & j < 1\\
        F(j{-}1), & j \ge 1\\

        \underset{0}{\max}~\underset{1}{\min}

            \{
            \underset{(i, j) \in A}{F(i{-}1)_0} + (j{-}i{+}1), F(i{-}1)_1{+}1 
            \}
        
        , & j \ge 1
        
        \end{array}\right\} 
\end{equation}
\end{scriptsize}
\normalsize
We define a recurrence to the disjoint interval covering problem in Equation~\ref{eq:1}. This recurrence posits that the segmentation that maximally covers the word is either the solution for the current word minus the ending character, or the max-covering, min-morpheme solution utilizing all morphemes that have a right boundary index equal to the index of the end of the word. With proper memoization, it is evident that for a word of size $|v|$, there are $|v|$ subproblems to solve. In addition, because each interval's right boundary corresponds to the word size, each interval is iterated over a constant number of times. As such, for word $v$, the total, memoized complexity of this segmentation is $\mathcal{O}(v + |A_v|)$ where $A_v$ indicates the pre-segmentation morphemes that are substrings of word $v$, making our overall framework of linear complexity -- $\mathcal{O}(V)$.

\begin{algorithm2e}
\caption{DP Parsimonious Segmentation (DP)}
\label{alg:subword_partitioning}
\Indm
\small
       \KwIn{Word $v$, morpheme Intervals $A_v$} 
       \KwOut{Optimal segmentation $S$}
\Indp
       \BlankLine
      $ \text{n}[0]\gets0;$
        $ \text{c}[0]\gets0;$
        $  \text{p}[0]\gets\text{null};$ \\
	\For{$j:=1 \text{ to } N_v$}{
		num $\gets$ n[j{-}1];
		cov$\gets$c[j{-}1];
		pair $\gets$ p[j{-}1];\\
		\For{$(i, j) \in A_v$}{
			cov$' \gets$ \text{c}[i{-}1]{+}(j{-}i{+}1)\\
			num$' \gets  \text{n}[i{-}1]{+}1$\\
			\If{\text{cov}$'$ $>$ cov}
			{
				\text{cov} $\gets$ cov$'$;
				\text{num} $\gets$ num$'$;\\
				\text{pair} $\gets (i,j)$;\\
			}
			\If{cov$'$=cov  $\wedge$ num$'$$<$num}
			{
				\text{num} $\gets \text{num}'$;~
				\text{pair} $\gets (i,j)$\\
			}
		   }
		   n[j] $\gets$ num;~
		   c[j] $\gets$ cov;~
		   p[j] $\gets$ pair;\\
	}
	  return p
\end{algorithm2e}

\normalsize

Algorithm~\ref{alg:subword_partitioning} presents the morpheme segmentation algorithm. The algorithm takes as input a  word and a collection of intervals corresponding to index boundaries of candidate morphemes within the word. It then proceeds to select a set of intervals that maximally cover the word while utilizing the fewest number of intervals. Solutions to subproblems are memoized as to avoid repeated computation. While the algorithm returns a memoization list of best segmentations that terminate at each index, proper backstracking can construct all possible parsimonious segmentations. In the next subsection we demonstrate how to select among equally-parsimonious segmentations.

\subsubsection{Maximum Likelihood Scoring}
While Algorithm~\ref{alg:subword_partitioning} identifies the most parsimonious segmentation, the algorithm often returns many segmentations with equal parsimony. As such, after applying Algorithm~\ref{alg:subword_partitioning}, maximum likelihood is used to select \textit{the most likely segmentation} among these candidate segmentations.

Given the previous counts of candidate morphemes obtained, it is simple to compute the most likely segmentation among the candidate set of parsimonious segmentations given an independence assumption. 
Given a segmentation (partition of morphemes) over word $v$, $\mathcal{G}_v$, one can calculate the likelihood over the partition:
\begin{equation*}
\mathcal{L}(\mathcal{G}_v) = \prod_{m\in\mathcal{G}_v} \mathsf{P}(m)  \\
\propto \prod_{m\in\mathcal{G}_v} f(m)
\end{equation*}
The independence assumption yields and discarding the normalization yields a simple product over each morpheme count $f(m)$ in the partition. This follows as all parsimonious partitions have the same number of morphemes and as such, the normalization constants for the probabilities should be the same for all parsimonious segments. One additional important constraint we place on our most-likely partition is that, during training and learning of the morpheme vocabulary, the most-likely partition cannot have any morphemes that occur only once in the vocabulary. That is: $\forall m\in \mathcal{G}_v: f(m) > 1$. This ensures that each learned morpheme is shared at least with another word. As seen in Example~\ref{ex:likely}, the largest product of counts is selected as the best segmentation. By applying the restriction that all morphemes must be shared at least once, MorphMine filters poor segmentations such as ``\textit{incompletenes} + \textit{s}'' where the morpheme ``\textit{incompletenes}'' only appears once in the vocabulary. 

\begin{example}[Most-likely Segmentation]
\label{eg:likely}
Most likely segmentation from candidates.
\begin{center}
\begin{tabular}{ llll}
  \hline
 \textbf{\emph{Segmentation}} & \textbf{\emph{$f(m_1)$}}  & \textbf{\emph{$f(m_2)$}} & \emph{Likelihood Score}\\
  \hline
  \hline
  {[\textbf{incompletenes}]} + {[\textbf{s}]} & 1 & 2072 & 2072 \\
  \hline
{[\textbf{incomplete}]} +  {[\textbf{ness}]} & 4 & 115 & 660 \\
\hline
  \rowcolor{gray!30}{[\textbf{in}]}  + {[\textbf{completeness}]} & 659 & 4 & 2636 \\
  \hline
    {[\textbf{incomp]}} +  {[\textbf{leteness}]} & 4 & 2 & 8\\
  \hline
\end{tabular}
\end{center}
\center
The highlighted row displays the most likely segmentation.
\label{ex:likely}
\end{example}
\vspace{0.2cm}
The mostly likely segmentation ``\textit{in} + \textit{completeness}'' is selected and in further steps, ``completness'' will be recursively decomposed into smaller morphemes ``complet" and ``ness" parsimoniously.

\subsection{Local Segmentation.}
Subsection~\ref{subsec:candidate}  introduced the concept of utilizing high-entropy boundaries to create a morpheme vocabulary, and Subsection~\ref{subsec:segmenting_words}  introduced an algorithm for segmenting words into morphemes based on the principle of parsimonious disjoint interval covering and tie-breaking with maximum likelihood. In this subsection we demonstrate a high-level overview on how to apply these two methods to hierarchically segment words into multi-granular morphemes.

\begin{algorithm2e}[ht]
\caption{Segmentation Algorithm (SEGMENT)}
\label{alg:hierarchical_segmentation}
\Indm
\small
       \KwIn{Word $v$, morpheme Vocabulary SW} 
       \KwOut{Set of morphemes of $v$}
\Indp
       \BlankLine
	output $\gets \{v\}$ \\
	$A_v \gets \{(i,j) \text{ for } v_i \ldots v_j \in SW$ and j{-}i $\ne$ $|v|\}$ \label{lst:line:interval} \\ 
	\If{$A_v = \varnothing$}
		{
		\textbf{return} output
		}
	segmented $\gets$ DP(w, $A_v$) \label{lst:line:parsimonious}\\
	
	\For{morpheme $\in$ segmented}{
		output $\cup$ SEGMENT(morpheme, SW)		
	}	
	  \textbf{return} output
\end{algorithm2e}
\normalsize

Following the steps from Subsection~\ref{subsec:candidate}, an initial morpheme vocabulary is created. Within the vocabulary, we differentiate between prefixes, suffixes, and root words. As seen in Algorithm~\ref{alg:hierarchical_segmentation}, Line~\ref{lst:line:interval}, each morpheme found in the input word is mapped to an interval indicating its boundary indices within the word with the condition that prefix intervals must start at the beginning of the word, suffix intervals must terminate at the end of the word, and root word intervals can be located at any position within the word. In addition, the complete word is not included (to ensure the word segments to smaller morphemes). The algorithm terminates if the word cannot be further segmented. Otherwise, the word is segmented with the dynamic programming parsimonious segmentation algorithm. Each morpheme is then treated as a word and recursively segmented; the collection of all morphemes from segmentation are output.

\subsection{Global Resegmentation}
The parsimonious segmentation selects the most-likely locally-consistent segmentation of a word. 
Yet because each word is segmented independently, a morpheme that is present in two different words may not be selected because parsimonious segmentation is performed on both words independently. 
To address this, after performing one segmentation we utilize the resultant segmentation to refine the morpheme counts and prune infrequent morphemes from the morpheme vocabulary. 
Using this refined morpheme vocabulary and a more accurate morpheme count estimation, each word is re-segmented. The resultant segmentation is not only performed with a smaller morpheme vocabulary, but also favors the morphemes that other words have selected in their own parsimonious segmentations, creating a global consistency for the overall vocabulary segmentation.

\vspace{0.2cm}
\begin{example}[Re-segmentation with refined counts.]
\label{eg:rsegmentation}
After one pass through the vocabulary and segmenting with parsimonious segmentation. Morpheme counts are re-computed using the resultant segmentations.
\begin{center}
\begin{tabular}{ llllll}
  \hline
 \textbf{\emph{Segmentation}} & \textbf{\emph{Counts$_1$}} &
 \textbf{\emph{ML$_1$}}&
 \textbf{\emph{Counts$_2$}} &
 \textbf{\emph{ML$_2$}}
 \\
  \hline
  \hline
  {[\textbf{bit}]} + {[\textbf{emporal}]} & 5,6 & \textbf{30}& 3,1 & 3 \\
  \hline
\rowcolor{gray!30}{[\textbf{bi}]} +  {[\textbf{temporal}]} & 4,6  & 24 & 3,5  & \textbf{15}  \\
  \hline
\end{tabular}
\end{center}
\vspace{0.2cm}

While initial counts and scores (Counts$_1$, ML$_1$) determine the locally optimal segmentation. After recomputing the morphemes post initial segmentation, refined counts and scores are computed (Counts$_2$, ML$_2$). The white row displays the initial best segmentation, while the grey row shows the best segmentation after refining morpheme counts.
\label{ex:resegmentation}
\end{example}
\vspace{0.2cm}

As seen in Example~\ref{ex:resegmentation}, in the first segmentation, the locally-consistent parsimonious segmentation favors the incorrectly segmented ``\textit{bit} + \textit{emporal}'' with likelihood 30 over ``\textit{bi} + \textit{temporal}'' with likelihood 24 as the likelihood is higher for the former.
After one round of segmentation, it is apparent that the morpheme ``\textit{emporal}'' was only selected once out of the possible six occurrences, while ``\textit{temporal}'' was selected five out of six word segmentations. Resegmentation with these refined counts helps choose the correct segmentation ``\textit{bi} + \textit{temporal}'' with likelihood score 15 over 3. With this refined segmentation, these morphemes can be used in the morpheme-enriched word embedding learning. 

\subsection{Morpheme-Enriched Word Embedding}
To efficiently utilize our mined morphemes to improve upon word embeddings, we modify the FastText model for word embeddings to use our extracted morphemes~\cite{bojanowski2017enriching} to enrich \textit{infrequent or out-of-vocabulary} words. As explained in the FastText paper, it is often the longest subword that captures the most semantic meaning. As such, we take each and every node in our word segmentation representing morphemes at every granularity and directly input the morphemes extracted from this layer to enrich each word in the vocabulary. 

We begin with a brief review of FastText, and then demonstrate integrating morphemes in place of the standard FastText enumerated subwords. First, we note that FastText utilizes the skip-gram objective with negative sampling yielding the following objective (for simplicity, $\ell(x) =  \log(1 + \text{exp}({-x}))$):
\begin{equation*}
	\label{eq:negativelog}
	\sum_{x=1}^W \Big [ \sum_{c\in \mathcal{C}_x} \ell(s(w_x, w_c)) +  \sum_{t\in\mathcal{N}_{x,c}} \ell(-s(w_x,t))\Big ],
\end{equation*}
where $w_x$ is the $x^{th}$ word in the corpus, $\mathcal{C}_x$ denotes the set of context words within a window of word $w_x$, and  $\mathcal{N}_{x,c}$ denotes the set of negative examples sampled from the vocabulary. The scoring function is then adapted to incorporate morpheme information as
$
    s(w_x, w_c) = \sum_{m\in w_x}\mathbf{z}_m^\intercal \mathbf{v}_c
$
where each $\mathbf{z}_m$ denotes a morpheme embedding vector and the scoring function is a summation over morpheme embedding vectors in a dot-product with the context word vector. While FastText incorporates all contiguous substrings of lengths three to seven as morphemes in the scoring function, we posit that many of these morphemes are semantically not meaningful and, as such, degrade the overall quality of the learned embeddings. We claim that directly incorporating meaningful morphemes extracted by MorphMine for each word and summing over each morpheme's embedding results in higher quality distributed representations. 
\section{Related Work}
\label{sec:related_works}
In morphological analysis, predictability has been suggested for detecting morpheme structure. An early quantitative metric proposed was the number of different variations of morphemes following a morpheme sequence whereby a high number of variations indicates a morpheme boundary~\cite{harris1970phoneme}. While this work provided influential insight into useful metrics for morpheme-detection, the main objective was developing a scoring function for identifying candidate morphemes, not segmentation. Following this line of work, were methods to identify frequent morphemes and affixes~\cite{hafer1974word,dejean1998morphemes,saffran1996word}. These methods identify a high-precision but low-recall subset of morphemes. Similarity measures have been proposed for detecting affixes by comparing words and identifying similar and dissimilar parts. These methods utilize a variety of techniques including edge-alignment, adding words and their reverse to tries~\cite{neuvel2002unsupervised,schone2001knowledge}.
Unfortunately, these methods can only identify prefix and suffix morphemes, ignoring morphemes. One model segments words by applying the minimum description length principle to minimize the vocabulary while maintaining the likelihood of the corpus data~\cite{creutz2002unsupervised,sak2010morphology}. 
Other fixed-vocabulary methods apply a unigram language model approach to identifying morphemes (also called wordpieces) and has been successfully applied to a variety of NLP tasks~\cite{wu2016google,schuster2012japanese}. Similarly, the byte-pair compression algorithm has been used to identify morphemes for neural machine translation tasks~\cite{sennrichneural}. 

To address data-sparsity when learning word embeddings, some methods apply a factored neural language model where words are represented as a set of features including morpheme information~\cite{alexandrescu2006factored}. Other methods add morphological similarity features into a neural network along with the context features~\cite{cui2015knet,qiu2014co}. Other methods take morphologically annotated data and train log-bilinear models to jointly predict context words and morphological tags~\cite{cotterell2015morphological}. The method we utilize for our embeddings is FastText~\cite{bojanowski2017enriching}. 
While FastText utilizes all the possible character $n$-grams up to certain length for enrichment, we only utilize high-quality morphemes in MorphMine. Finally, many methods have utilized characters as the base unit for embedding. Some approaches treat each word as a sequence of characters and apply RNNs or convolutional networks~\cite{bojanowski2015alternative,sutskever2011generating,kim2016character}.
 
\section{Experimental Results}

\label{sec:exp}
We introduce the datasets used and methods for comparison. We then evaluate our method on a morpheme segmentation task, a variety of embedding tasks, and a downstream language modeling task.

\smallskip
\noindent
\textbf{Datasets} 
\begin{itemize}
\item \textbf{English, German, and Turkish Vocabularies and Segmentations}. This dataset consists of three vocabulary lists in English, German, and Turkish with 156K, 290K and 90K unique vocabulary words, respectively. Each list is accompanied by approximately 1500 ground-truth segmentations consisting of a vocabulary word and its segmentation into constituent morphemes. These ground-truth segmentations were annotated as part of the MorphoChallenge~\cite{kurimo2010morpho}.

\item \textbf{English, German, and Turkish Wikipedia Corpora}. This dataset consists of three subsets of Wikipedia for English, German, and Turkish Wikipedia and consisting of 116M, 162M, and 52M tokens. These corpora are used for training unsupervised word embeddings and for training a language model.

\item \textbf{English, German, and Turkish Word Similarity Pairs}. This dataset consists of collections of annotated word-similarity pairs in three languages. For English, we evaluate on the WS-353 data, a collection of $353$ pairs of English words that have been assigned similarity ratings by human annotators, SimLex, a collection of $999$ word pairs annotated via Amazon Mechanical Turk, and finally the Stanford Rare Words similarity set (RW) consisting of $2034$ rare word pairs.~\cite{finkelstein2002placing,hill2015simlex,luong2013better}. For German, we operate on canonical translations of the the WS-353 and SimLex datasets~\cite{barzegar2018semr}. For Turkish we evaluate on the AnlamVer word similarity dataset consisting of $500$ word-pairs annotated by $12$ human annotators~\cite{ercan2018anlamver}.

\item \textbf{English, German, and Turkish Word Analogies}. Collections of annotated word analogies in three languages. For English, we evaluate on the Google analogy dataset consisting of 19544 analogy question pairs where $8,869$ are semantic and $10,675$ syntactic (i.e. morphological) questions.~\cite{mikolov2013efficient}. For German, we operate on the German translation of the English Google analogy dataset~\cite{koper2015multilingual}. For Turkish, counterparts of the Google analogy question set was created and contains over 2K analogy tasks.
\end{itemize}

The vocabulary lists and gold-standard segmentations are used to evaluate each method's ability to extract human-verified morphemes in an unsupervised manner. The human-curated word analogies and word similarity pairs help verify the effect of incorporating various morphemes in the unsupervised word embedding process. Finally, the Wikipedia corpora subsets are used to train the morpheme-enriched word embeddings and evaluate the benefit of morpheme enrichment on a downstream language modeling task.

\begin{table*}
\begin{center}
\begin{tabular}{ | c || ccc || ccc|| ccc |} \hline
\textbf{Dataset} & \multicolumn{3}{c||}{\textbf{English}} &  \multicolumn{3}{|c||}{\textbf{German}} & \multicolumn{3}{|c|}{\textbf{Turkish}} \\ \hline \hline
\textbf{Method} & P & R & F1 & P & R & F1
& P & R  & F1 \\ \hline\hline
BPE  & 0.5527 & 0.3989 &  0.4634
& 0.5637 & 0.4131 & 0.4768
& 0.7626  & 0.2808 & 0.4104 \\
ULM & 0.7473 & 0.5992 &  0.6651
& 0.5827 & 0.5040 & 0.5405
& \textbf{0.8731}  & 0.3216 & 0.4701 \\
Morfessor & 0.7537	 & 0.6513 &  0.6987
& \textbf{0.6803} & 0.5616& 0.6153
& 0.69104 & 0.3710 & 0.4828\\ 
\hline
MorphMine-NoRefine & 0.8255 & 0.6503 & 0.7275  
& 0.5717 & \textbf{0.7520} & 0.6399
&0.5894 & 0.5024 & \textbf{0.5424} \\
MorphMine & \textbf{0.8345} & \textbf{0.6977} & \textbf{0.7600}
& 0.6014 & 0.7373 & \textbf{0.6624}
& 0.5341 & \textbf{0.5497} & 0.5417\\\hline
\end{tabular}
\vspace{1mm}
\caption{Morpheme Segmentation Performance.}
\label{table:segmentation}
\end{center}
\end{table*}
\noindent
\textbf{Baselines}
\newline
As a baseline for segmentation, we utilize a unigram language model segmentation of ``word-pieces" and byte-pair encoding segmentation as described in the related work~\cite{schuster2012japanese,wu2016google}. We also compare against a state-of-the-art unsupervised morpheme segmentation tool Morfessor~\cite{creutz2002unsupervised}. Finally, we compare against a variant of MorphMine that forgoes global consistency whereby each word is re-segmented after recomputing morpheme counts after the initial segmentation. 

For baseline embedding methods, we utilize FastText, a proposed variation of the Skip-Gram objective that utilize subword-level information, and modify FastText to incorporate each method's segmentations to enrich word embedding. We enrich FastText with each of the morpheme segmentation baselines to compare against MorphMine enriched emebeddings. With no morpheme enrichment, FastText formulation means that it reduces to Word2Vec which we also compare against.

\subsection{Subword Extraction Accuracy}
We evaluate each morpheme segmentation algorithm at identifying human-annotated segmentations in three languages: English, Turkish, and German. We report precision, recall and F1 scores for each method. When evaluating, true-positives are indicated with a valid exact match between the extracted morpheme and the gold-standard.

In Table~\ref{table:segmentation}, we report the performance of each segmentor at successfully extracting human-annotated morphemes. As both BytePair Encoding and Unigram-LM require a morpheme vocabulary size parameter, for these methods, we perform a parameter sweep and report results from the highest performing run. Across all three languages, variants of MorphMine outperform with respect to F1 score. Further analysis shows this is primarily due to a higher recall. In comparison to MorphMine without global refinement, we see that implementing global refinement generally improves performance as seen in English and German and in the case of Turkish, performance between the MorphMine variants were overall comparable.
\subsection{Word Similarity Task}
We evaluate the embeddings on a word similarity task. The ground truth data consists of pairs of words and a human-annotated similarity score averaged across all human evaluations. The scores are computed via the cosine similarity between each word's vector representation and results are quantified through Spearman's rank correlation coefficient between the gold standard and the cosine similarity score. To evaluate performance of the morpheme-based embeddings to infer OOV words, we evaluate similarity on an English rare-words similarity dataset.

\begin{table*}
    \renewcommand{\arraystretch}{1.2}
\begin{center}
    \centering
\begin{tabular}{ | c || cccc || cc|| c |} \hline
\textbf{Method} & \multicolumn{4}{c||}{\textbf{English}} &  \multicolumn{2}{c||}{\textbf{German}} & \multicolumn{1}{c|}{\textbf{Turkish}} \\ \hline \hline
\textbf{Dataset} & \textbf{WS-353} & \textbf{SimLex} & \textbf{RW-Frequent} & \textbf{RW-OOV} & \textbf{WS-353} & \textbf{SimLex} & \textbf{AnlamVer}  \\ \hline
SkipGram & 0.72  & \textbf{0.28} & 0.36 & --  & 0.58 & 0.26 & 0.45 \\
BPE  & 0.72 & \textbf{0.28} & 0.41 & 0.33 &  0.59 & \textbf{0.28} & 0.47 \\
ULM & \textbf{0.74} & \textbf{0.28} & 0.41 & 0.35 &  0.60 & \textbf{0.28} & 0.47  \\
FastText & 0.70 & 0.26 & 0.34 & 0.32 & 0.58 & 0.26 & 0.46 \\
Morfessor & \textbf{0.74} & \textbf{0.28} & 0.44 & 0.35 & 0.60 & \textbf{0.28} & 0.48 \\ 
MorphMine & \textbf{0.74} & \textbf{0.28} & \textbf{0.46} & \textbf{0.42} & \textbf{0.62} & \textbf{0.28} & \textbf{0.49}\\\hline
\end{tabular}
\vspace{1mm}
\captionof{table}{Multilingual word similarity.}
\label{table:similarity}


\end{center}
\end{table*}
As seen in Table~\ref{table:similarity}, subword-based methods that utilize morpheme and subword level information outperform SkipGram that forgoes any. Additionally, methods that discriminately generate these morphemes outperform FastText that indiscriminately generate all subwords. Finally, while most subword enriched embeddings perform well on word similarity, MorphMine shines particularly in the similarity task on rare words where it outperforms all baselines. This is likely because MorphMine generates morphemes at multiple granularity which is more likely semantically link a rare word to a frequent word via a semantically-meaningful morpheme. This performance gap is even higher for out-of-vocabulary words were MorphMine significantly outperforms all other baselines at the word similarity task.

\subsection{Word Analogy Task}
We next evaluate on a word analogy task of the form ``$A$ is to $B$'' as ``$C$ is to $D$'', where $D$ is predicted from the vocabulary based on its embedding vector. We use analogy datasets used in previous literature for English, German, and Turkish embedding evaluation~\cite{mikolov2013efficient,koper2015multilingual,sahin2016classification}. 

\begin{table}[H]
\centering
\begin{tabular}{ | c || cc || cc|| c |} \hline
\textbf{Dataset} & \multicolumn{2}{c||}{\textbf{English}} &  \multicolumn{2}{c||}{\textbf{German}} & \multicolumn{1}{c|}{\textbf{Turkish}} \\ \hline \hline
\textbf{Method} & \textbf{Sem} & \textbf{Syn} & \textbf{Sem} & \textbf{Syn} & \textbf{Sem+Syn} \\ \hline
SkipGram & \textbf{68} & 65 & \textbf{63}  & 46 & 41 \\
BPE  & 65 & 68 & 61  & 50 & 43 \\
ULM & 67 & 70 & 62 & 51  & 43  \\
FastText & 52 & 75  & 59 & \textbf{53} & 43 \\
Morfessor & 64 & 75  & 61 & 52 & 44 \\ 
MorphMine & 67 &  \textbf{78} & 61 & \textbf{53} & \textbf{47}\\
\hline
\end{tabular}
\vspace{1mm}
\caption{Word analogies.}
\label{table:analogies}
\end{table}

As seen in Table~\ref{table:analogies}, embeddings that utilize subword information perform better at syntactic analogies than SkipGram word embeddings without subword information. This does not extend to semantic analogies whereby utilizing subword-information seems to cause a deterioration in performance. This is intuitive as words without valid morphemes learn noisy embeddings when false morphemes are identified and used to enrich their representation. This is seen in the performance gap between FastText and SkipGram on semantic analogies whereby FastText's large number indiscriminate subwords degrades the quality of the final embedding. This degradation is mitigated by utilizing more-refined morpheme methods such as BytePair Encoding, Unigram-LM, Morfessor, and MorphMine. Overall, embeddings enriched with MorphMine morphemes demonstrate superior syntactic performance to all baselines while demonstrating comparable semantic performance to SkipGram. This supports our intuition that utilizing more subwords is useful, but only when they are of high quality; indiscriminately generating all enumerations of subword degrades quality.
%

\subsection{Language Modeling Perplexity}
As recent embedding evaluations have stressed the importance of evaluating embeddings not only on artificial tasks such as word similarities but also on downstream tasks, we evaluate on a downstream language modeling task~\cite{faruqui2016problems}.
We generate a language model with embedding vectors from the Wikipedia corpora and then evaluate by computing the perplexity on a held-out portion of the corpus \textit{unseen in both the embedding phase and modeling phase}. We use an LSTM with two hidden layers, $600$ hidden units per layer regularized with dropout with $0.2$ probability, unrolled for 35 steps, and $20$ batch size. Parameters are learned using Adagrad with a gradient clipping of $1$ for 10 epochs. Each instance is trained on $80\%$ of the data with a $10\%$ test and $10\%$ validation set.

\begin{table}[H]
\centering
\begin{tabular}{ | c || c || c|| c |} \hline
\textbf{Dataset} & \multicolumn{1}{c||}{\textbf{English}} &  \multicolumn{1}{c||}{\textbf{German}} & \multicolumn{1}{c|}{\textbf{Turkish}} \\ \hline \hline
\textbf{Method} & \textbf{Perplexity} & \textbf{Perplexity} & \textbf{Perplexity} \\ \hline
SkipGram & 159 & 381 & 996  \\
BPE  & 157 & 375 & 955  \\
ULM & 157 & 372 & 952  \\
FastText & 158 & 376  & 972 \\
Morfessor & 155 & 370  & 948 \\ 
MorphMine & \textbf{154} &  \textbf{367} & \textbf{940} \\\hline
\end{tabular}
\vspace{1mm}
\caption{Language modeling task}
\label{tab:ppl}
\end{table}
\begin{table*}
    \renewcommand{\arraystretch}{1.2}
\begin{center}
\begin{tabular}{|c||c||c||c||c|}
\hline
\textbf{Word}  & \textbf{BPE} & \textbf{ULM} & \textbf{Morfessor} & \textbf{MorphMine} \\
\hline
\hline
 vandalism  & van + dal + ism & van + dal + ism & van + dal + ism & vandal + ism\\
\hline
 truncate  & trun + cate & trun + cate & truncate & truncat + e\\
\hline
 truncated  & trun + cat + ed & trun + cat + ed & truncat + ed & truncat + ed\\
 \hline
 truncating & trun + cat + ing & trun + cat + ing & truncat + ing & truncat + ing \\
  \hline
 troubleshooting & trouble + shoot + ing & trouble + shoot + ing & trouble + shoot + ing & \makecell{troubleshoot + ing + \\ trouble + shoot} \\
\hline
\end{tabular}
\end{center}
\vspace{-1mm}
\caption{Select segmentations from different subword segmentation algorithms.}
\label{tab:segments}
\end{table*}

The results are summarized in Table~\ref{tab:ppl}. Because experiments have minimal data cleaning do not drop infrequent or OOV words, the resulting perplexity is relatively higher than cleaned-datasets but directly comparable among the differing methods~\cite{bojanowski2017enriching}. We observe that across all segmentation-based morpheme-enriched embeddings perform better in language modeling over traditional skip-gram. In contrast, FastText's indiscriminate enumeration of all possible morphemes appears to perform much poorer in this task. Finally, MorphMine outperforms the other morpheme enriched baselines. This may be due to MorphMine utilizing morphemes of mutliple granularity which closely capture semantic meaning of rare and OOV words at the largest granularity.

\subsection{Segmentation Case Study}

In Table~\ref{tab:segments}, we present hand-selected segmentations. Unlike other methods, MorphMine identifies large morphemes shared among words in the vocabulary in addition to the more frequent smaller morphemes. For example, the words \textit{``truncate", ``truncated", ``truncating"} all share a common root, but all methods except for MorphMine are reluctant to identify \textit{``truncat"} as a valid morpheme by removing `e' from \textit{truncate}. As such, all other methods fail semantically link these three words. Additionally, for `\textit{vandalism}', most methods attempt to recognize ``\textit{van}'' as a morpheme as it is a valid word, while MorphMine's parsimony criterion merges this into ``\textit{vandal}'', which although not present in the vocabulary, is a valid word. 
Finally, given words such as \textit{``troubleshooting"}, MorphMine's segmentation at multiple granularities captures ``\textit{troubleshoot}'', which all other methods further decompose, losing much semantic meaning.

\subsection{Scalability}
From a high-level perspective, MorphMine consists of two separate steps: (1) mining and learning a high-quality candidate morpheme set from an input vocabulary and (2) utilizing the learned model to segment each word into morphemes. We can empirically estimate the expected runtime of each step of MorphMine by analyzing runtime as a function of input size.

\begin{figure}
\centering
\includegraphics[width=0.47\textwidth]{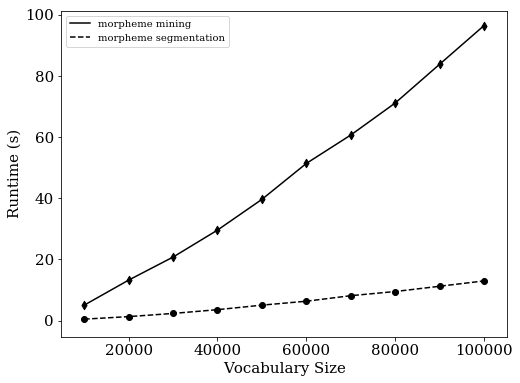}
\caption{Decomposition of morpheme segmentation algorithm into unsupervised morpheme mining then vocabulary segmentation.}
\label{fig:decomp}
\end{figure}

As seen in Figure~\ref{fig:decomp}, mining the morpheme vocabulary appears to grow linearly with vocabulary size. We verify this by computing the coefficient of determination, $R^2$ to show how well a linear function fits the data. Morpheme mining and segmentation regressions yielded an $R^2$ of $0.989$ and $0.991$ respectively. This strongly suggests a linear relationship between input vocabulary size and runtime. As empirically Heap-Herdan's law has shown that vocabulary grows sublinearly in relation to corpus size, these results indicate that performing MorphMine segmentation on an input vocabulary as a preprocessing step adds negligible computational overhead~\cite{egghe2007untangling}.

\section{Conclusions}
In this study, we propose a pattern-mining method of segmenting vocabulary into smaller morphemes and demonstrate experimentally on three languages that the method recovers ground-truth morphemes beyond state-of-the-art. By integrating the morphemes in a popular subword-enriched embedding algorithm, we verify that semantically-meaningful morphemes at multiple granularity can benefit word embeddings as evidenced through superior performance on a word analogy and word similarity task. This is especially true for inferring embeddings for infrequent or out-of-vocabulary words. Finally, we demonstrate that enriching embeddings with high-quality morphemes improves language modeling as evidenced through better held-out perplexity on a language modeling task. 


\section{Acknowledgements}
Research was sponsored in part by U.S. Army Research Lab. under Cooperative Agreement No. W911NF-09-2-0053 (NSCTA), DARPA under Agreements No. W911NF-17-C-0099 and FA8750-19-2-1004, National Science Foundation IIS 16-18481, IIS 17-04532, and IIS-17-41317, DTRA HDTRA11810026, and grant 1U54GM114838 awarded by NIGMS through funds provided by the trans-NIH Big Data to Knowledge (BD2K) initiative (www.bd2k.nih.gov). Any opinions, findings, and conclusions or recommendations expressed in this document are those of the author(s) and should not be interpreted as the views of any U.S. Government. The U.S. Government is authorized to reproduce and distribute reprints for Government purposes notwithstanding any copyright notation hereon.
\balance

\bibliographystyle{plain}
\bibliography{icdm}

\end{document}